\documentclass[conference]{IEEEtran}
\IEEEoverridecommandlockouts
\usepackage[adjust]{cite}
\usepackage{amsmath,amssymb,amsfonts}
\usepackage{algorithmic}
\usepackage{graphicx}
\usepackage{textcomp}
\usepackage{xcolor}
\def\BibTeX{{\rm B\kern-.05em{\sc i\kern-.025em b}\kern-.08em
    T\kern-.1667em\lower.7ex\hbox{E}\kern-.125emX}}

\usepackage{stackrel}
\usepackage{multicol}
\usepackage{bold-extra}
\usepackage{xspace}
\usepackage{siunitx}
\DeclareSIUnit{\SIpi}{\ensuremath{\mathnormal{\pi}}}
\usepackage{tikz}
\usepackage{svg}
\usepackage{booktabs}
\usepackage[misc]{ifsym}

\makeatletter
\newcommand{\linebreakand}{%
  \end{@IEEEauthorhalign}
  \hfill\mbox{}\par
  \mbox{}\hfill\begin{@IEEEauthorhalign}
}
\makeatother

\newcommand{\argos}{ARGoS3\xspace}
\newcommand{\auto}{AutoMoDe\xspace}

\newcommand{\cho}{\texttt{Choc\-o\-late}\xspace}

\newcommand{\evsti}{\texttt{EvoStick}\xspace}

\newcommand{\Magg}{\textsc{Aggregation}\xspace}
\newcommand{\Mgri}{\textsc{Grid Exploration}\xspace}
\newcommand{\Mfor}{\textsc{Foraging}\xspace}

\newcommand{\epuck}{e-puck\xspace}
\newcommand{\epucks}{e-pucks\xspace}

\newcommand{\mercator}{Mercator\xspace}
\newcommand{\mercators}{Mercators\xspace}

\newcommand{\RM}[1]{RM\,#1\xspace}
\begin{document}

\title{Automatic off-line design of robot swarms: exploring the transferability of control software and design methods across different platforms
\thanks{\IEEEauthorrefmark{1}MK and DGR contributed equally to this work and should be recognized as co-first authors. Original software was implemented by MK. \cho and \evsti were developed by GF. The experiments were performed by MK with the assistance of DGR. The manuscript was drafted by DGR and revised by MK and MB. All authors contributed to the development of the ideas, read the manuscript, and provided comments. The research was directed by MB.  
Correspondence to mauro.birattari@ulb.be.

\smallskip
The project has received partial funding from the European Research Council (ERC) under the European Union’s Horizon 2020 research and innovation programme (DEMIURGE Project, grant agreement No 681872) and from Belgium’s Wallonia-Brussels Federation through the ARC Advanced project GbO-Guaranteed by Optimization.
The authors acknowledge support from the Belgian Fonds de la Recherche Scientifique – FNRS and from the Colombian Ministry of Science, Technology and Innovation – Minciencias.
}}

\author{\IEEEauthorblockN{Miquel Kegeleirs\IEEEauthorrefmark{1}}
\IEEEauthorblockA{\textit{IRIDIA, Université libre de Bruxelles} \\
Brussels, Belgium \\
miquel.kegeleirs@ulb.be}
\and
\IEEEauthorblockN{David Garzón Ramos\IEEEauthorrefmark{1}}
\IEEEauthorblockA{\textit{IRIDIA, Université libre de Bruxelles} \\
Brussels, Belgium \\
david.garzon.ramos@ulb.be}
\and
\IEEEauthorblockN{Lorenzo Garattoni}
\IEEEauthorblockA{\textit{Toyota Motor Europe} \\
Brussels, Belgium \\
lorenzo.garattoni@toyota-europe.com}
\linebreakand
\IEEEauthorblockN{Gianpiero Francesca}
\IEEEauthorblockA{\textit{Toyota Motor Europe} \\
Brussels, Belgium \\
gianpiero.francesca@toyota-europe.com}
\and
\IEEEauthorblockN{Mauro Birattari\textsuperscript{~\Letter}}
\IEEEauthorblockA{\textit{IRIDIA, Université libre de Bruxelles} \\
Brussels, Belgium \\
mauro.birattari@ulb.be} 
}

\maketitle

\begin{abstract}
Automatic off-line design is an attractive approach to implementing robot swarms.
In this approach, a designer specifies a mission for the swarm, and an optimization process generates suitable control software for the individual robots through computer-based simulations.
Most relevant literature has focused on effectively transferring control software from simulation to physical robots.
For the first time, we investigate (i)~whether control software generated via automatic design is transferable across robot platforms and (ii)~whether the design methods that generate such control software are themselves transferable.
We experiment with two ground mobile platforms with equivalent capabilities.
Our measure of transferability is based on the performance drop observed when control software and/or design methods are ported from one platform to another.
Results indicate that while the control software generated via automatic design is transferable in some cases, better performance can be achieved when a transferable method is directly applied to the new platform. 
\end{abstract}


\section{Introduction} 

% 0.25 page Automatic design
%
A robot swarm~\cite{Sah2005sab,Ben2005sab} is a highly redundant group of robots that operates autonomously without relying on centralized control or external infrastructure.
Instead, swarm individuals rely on local sensing and communication to self-organize~\cite{DorBirBra2014SCHOLAR}. 
By acting collectively, robots can accomplish tasks that they could not accomplish individually~\cite{DorTheTri2021PIEEE}. 

Designing the collective behavior of a swarm is particularly challenging.
No universally applicable methodology exists for developing the control software of the individual robots so that a desired collective behavior emerges~\cite{BraFerBirDor2013SI}.
Typically, designers manually refine control software until the desired collective behavior is obtained.
This trial-and-error design process is costly, time-consuming, and does not guarantee reproducible or transferable results.
Automatic off-line design~\cite{BirLigBoz-etal2019FRAI} is an appealing alternative.
In this approach, the problem of designing control software for individual robots is re-formulated as an optimization problem.
Given mission specifications and a platform description, an optimization algorithm searches for suitable control software for the robots.
The automatic design process aims to produce control software that maximizes swarm performance in the mission at hand, according to a mission-specific performance metric provided as part of the mission's formal specification.

\emph{Neuroevolution}~\cite{Tri2008book} is the traditional approach to the automatic off-line design of robot swarms. 
In this approach, control software takes the form of an artificial neural network with parameters tuned via artificial evolution.
More recently, \emph{automatic modular design}~\cite{FraBraBru-etal2014SI} has been proposed as an alternative to neuroevolution.
In modular design, an optimization algorithm selects and tunes predefined software modules into a specific control architecture (e.g., finite-state machines~\cite{FraBraBru-etal2014SI} or behavior trees~\cite{LigKucBozBir2020PEERJCS}).
Both in neuroevolution and the modular approach, the design process is conducted via computer-based simulations.
The resulting control software is then transferred to physical robots and assessed in the target environment.

Control software produced via automatic off-line design suffers from the \emph{reality gap}~\cite{JakHusHar1995ecal}.
Unavoidable differences between the design environment (simulation) and the deployment environment (physical robots) can cause a performance drop in the swarm~\cite{FraBraBru-etal2014SI,FraBraBru-etal2015SI,LigBir2020SI,HasLigRudBir2021NATUCOM,Lig2023phd}.
When comparing design methods, the smaller the performance drop, the greater a method's ability to cross the reality gap.

Building on this idea, we study the \emph{transferability} of control software and automatic design methods across different robot platforms.
We contend that the process of transferring control software across different platforms gives rise to challenges akin to those presented by the reality gap.
When discussing transferability, we consider cases where the control software is designed for one platform and then deployed on another, or when an automatic design method conceived for one platform is applied to another. 
This is reminiscent of the reality gap problem, which emerges when designing control software in simulation for execution on real robots.

In this preliminary study, we first investigate the conditions under which modular and neuroevolutionary design methods conceived for a given robot platform can produce control software for a different one.
Then, we investigate whether control software produced via automatic design can be directly transferred from one platform to another.
The two platforms considered are ground robots endowed with equivalent sensing and actuation capabilities.

\section{Experiments}

\begin{figure}
    \centering
    \includegraphics[width=\linewidth]{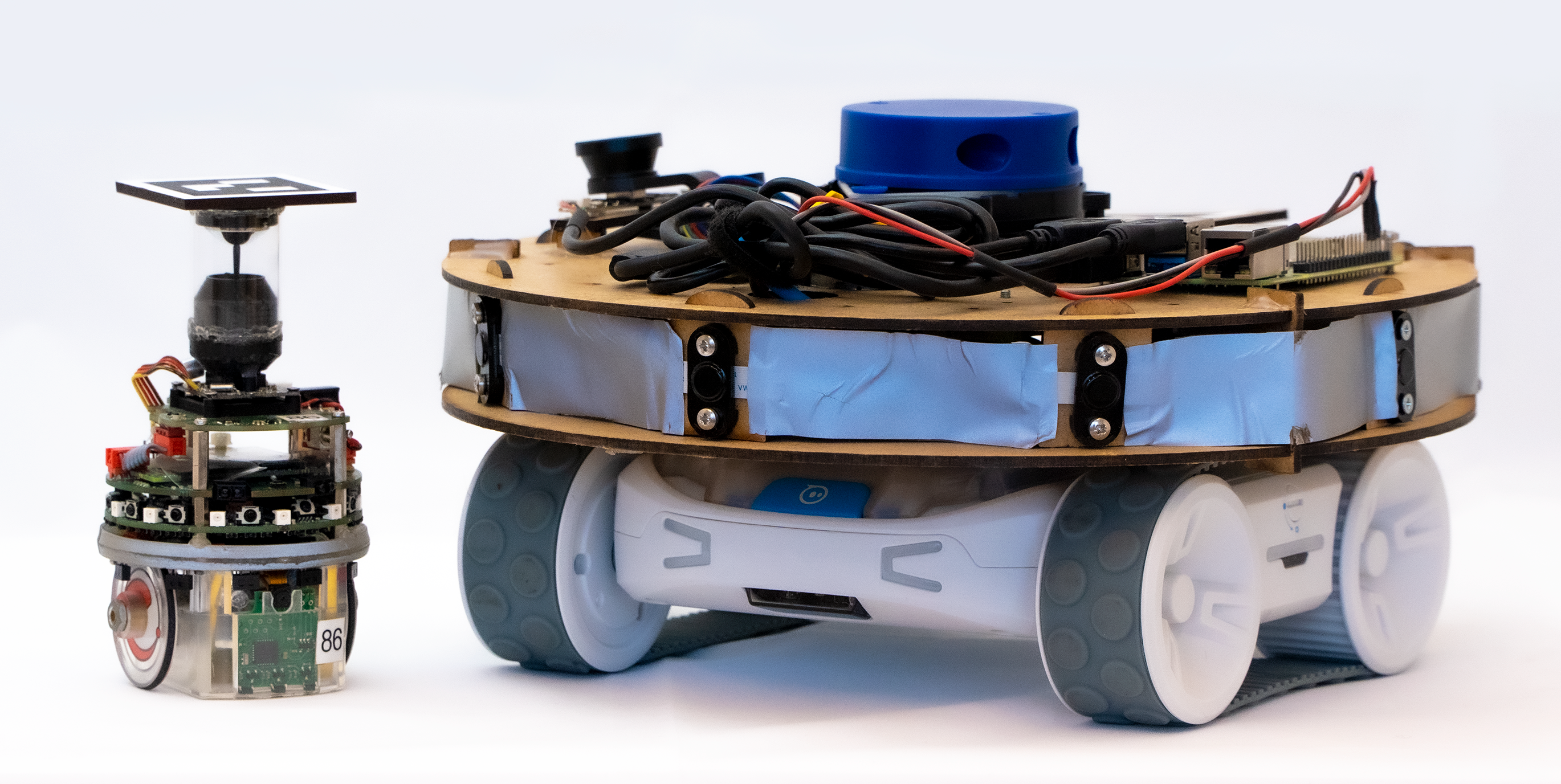}
    \caption{From left to right, \epuck and \mercator robots.}
    \label{fig:robots}
\end{figure}

%Figure~\ref{fig:experiment} illustrates the experimental setup.
%
We consider two swarms:
one comprising three \epuck~\cite{MonBonRae-etal2009arsc,GarFraBru-etal2015techrep} robots, and the other, three \mercator~\cite{KegTodGar-etal2022techrep} robots.
We can formally describe the sensing and actuation capabilities of the two platforms with the same reference model~\RM{1.2}~\cite{HasLigFra-etal2018techrep}---see Table~\ref{tab:RM12}---which defines the inputs and outputs on which control software operates~\cite{FraBraBru-etal2014SI}.
This is key to enabling the transferability between the two platforms.
The \epuck and \mercator (Figure~\ref{fig:robots}) differ in size, with the \epuck being roughly one-third the size of the \mercator.
They also differ in linear speed and sensor range.
However, their speed/size and sensor-range/size ratios are approximately the same.

\begin{table}
	
	\caption{Reference model \RM{1.2}. Input and output values of \epucks'~(EP) and \mercators'~(ME) control software.}
    \label{tab:RM12}
	\centering
	\begin{tabular}{>{\raggedright}p{0.8cm}>{\raggedright}p{2.0cm}>{\raggedright}p{2.0cm}>{\raggedright}p{1.9cm}}
		\toprule
		\textbf{Input} & \textbf{Value--EP} & \textbf{Value--ME} & \textbf{Description} \tabularnewline
		\midrule
		$\mathit{prox}$ & $(\SIrange{0}{1}{};\SIrange{-1}{1}{\pi})$ & $(\SIrange{0}{1}{};\SIrange{-1}{1}{\pi})$ & proximity vector \tabularnewline
        $\mathit{light}$ & $(\SIrange{0}{1}{};\SIrange{-1}{1}{\pi})$ & $(\SIrange{0}{1}{};\SIrange{-1}{1}{\pi})$ & light vector \tabularnewline
		$\mathit{gnd}$ & $\{\mathit{b},\mathit{g},\mathit{w}\}$ & $\{\mathit{b},\mathit{g},\mathit{w}\}$ & ground reading \tabularnewline
        $\mathit{n}$ & $[0,2]$ & $[0,2]$ & no. neighbors \tabularnewline
		$\mathit{V}$ & $(\SIrange{0}{1}{};\SIrange{-1}{1}{\pi})$ & $(\SIrange{0}{1}{};\SIrange{-1}{1}{\pi})$ & neighbors vector \tabularnewline
		\midrule
		\textbf{Output} & \textbf{Value--EP} & \textbf{Value--ME} & \textbf{Description} \tabularnewline
		\midrule
		$v_{k\in\{l,r\}}$ & $\SIrange{-10}{10}{\cm\per\second}$ & $\SIrange{-30}{30}{\cm\per\second}$ & target velocity \tabularnewline
		\bottomrule 
	\end{tabular}\\
    \smallskip
	\begin{tabular}{cccc}
		\multicolumn{1}{c}{\footnotesize Period of the control cycle: $\SI{0.1}{\second}$.}
	\end{tabular}
\end{table}

\begin{figure}
    \centering  
    \setlength{\unitlength}{1cm}
    \begin{picture}(\linewidth,0.63\linewidth)(0.1,0)
    \put(0.075\linewidth,0){\includegraphics[width=0.925\linewidth]{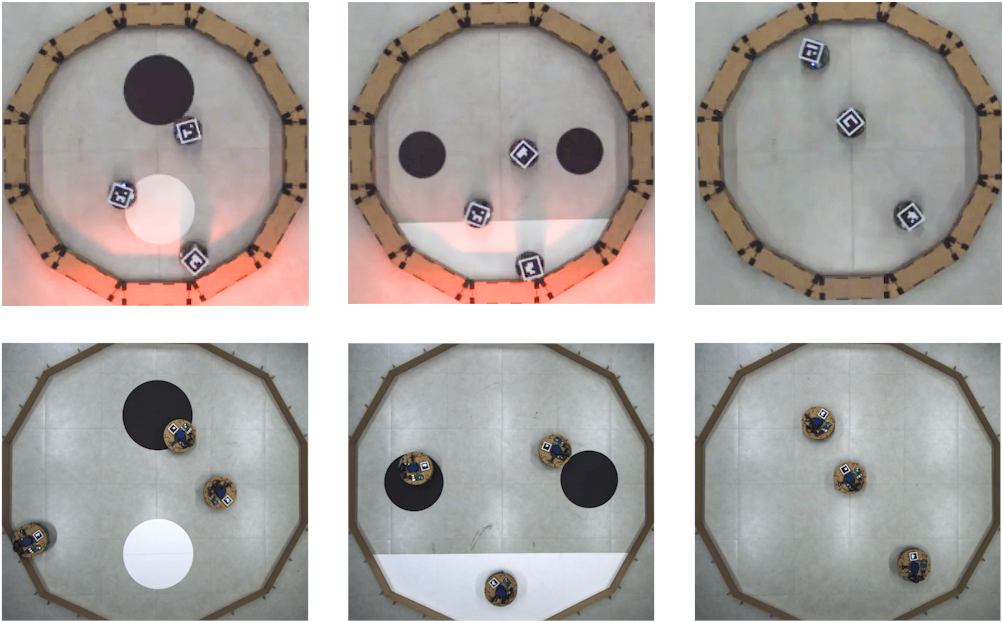}}
    \thicklines
    \put(0.025\linewidth,0.035\linewidth){\rotatebox[origin=l]{90}{\mercators}}
    \put(0.025\linewidth,0.36\linewidth){\rotatebox[origin=l]{90}{\epucks}}
    \put(0.105\linewidth,0.6\linewidth){\makebox(0,0)[l]{\  \footnotesize \Magg}}
    \put(0.45\linewidth,0.6\linewidth){\makebox(0,0)[l]{\  \footnotesize \Mfor}}
    \put(0.71\linewidth,0.6\linewidth){\makebox(0,0)[l]{\  \footnotesize \Mgri}}
    \put(0.79\linewidth,0.37\linewidth){\makebox(0,0)[l]{\  \footnotesize \SI{30}{\cm}}}
    \put(0.79\linewidth,0.35\linewidth){\line(1,0){0.875}}
    \put(0.774\linewidth,0.06\linewidth){\makebox(0,0)[l]{\  \footnotesize \SI{100}{\cm}}}
    \put(0.79\linewidth,0.04\linewidth){\line(1,0){0.875}}
    \end{picture}
    \caption{Experimental scenarios in \Magg, \Mfor, and \Mgri. The workspace of the \epucks is about one-third the size of \mercators' workspace. The pictures show an example of the initial configuration at the beginning of each mission.}
    \label{fig:arenas}
\end{figure}
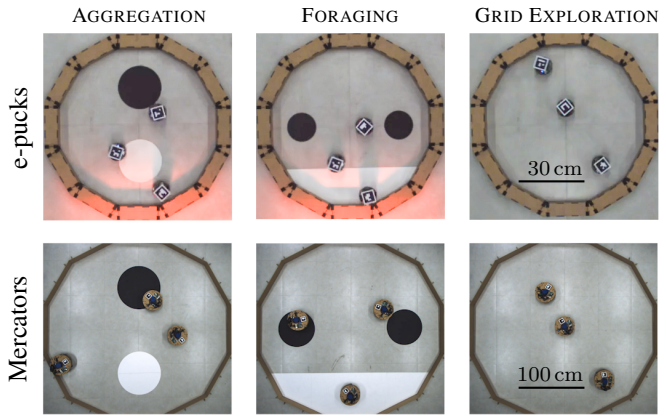

We design control software for \epucks and \mercators using two automatic methods originally conceived for the \epuck: \cho and \evsti.
\cho~\cite{FraBraBru-etal2015SI} is a modular design method from the \auto~\cite{BirLigFra2021admlsa} family and \evsti~\cite{FraBraBru-etal2014SI} is an implementation of the neuroevolutionary approach. 
We select these two methods because they have been largely used in the automatic design of collective behaviors for \epucks~\cite{BirLigFra2021admlsa}---both in simulation and reality.
Previous results confirm that they can generate control software for various missions, including aggregation, foraging, and coverage~\cite{FraBraBru-etal2014SI,FraBraBru-etal2015SI,LigBir2020SI,HasLigRudBir2021NATUCOM,SpaKegGarBir2020bnaic-post}.
In our experiments, we consider three missions: \Magg, \Mfor, and \Mgri---see Figure~\ref{fig:arenas}.
In \Magg, robots must aggregate on a black spot.
In \Mfor, robots must iteratively travel between small black spots and a larger white region.
In \Mgri, robots must explore the environment and continuously visit every cell of a grid.
We adjust the size of the environment according to the relative size of the robots:
the \epucks operate in an environment that is one-third the size of the \mercators' one.
The performance of the two swarms is computed using the same performance measures.
We produce a total of 120 instances of control software using \cho and \evsti---ten for each  platform, mission, and method.
We assess each instance once in simulation (expected performance) and once with physical robots (real performance).
We conduct simulations in \argos~\cite{PinTriOgr-etal2012SI}---a simulator widely used in swarm robotics research.

\section{Results and discussion}

When assessed in simulation, the control software produced by both \cho and \evsti, for both \epucks and \mercators, demonstrates meaningful behavior and effectively performs the missions.
When assessed in reality, the control software produced by \cho maintains satisfactory behavior on both platforms.
In contrast, control software produced by \evsti does not typically reproduce simulation results on either platform.
The differences between simulation and reality are known effects of the reality gap:
previous research has shown that design methods from the \auto family are more robust to the reality gap than those based on neuroevolution~\cite{LigBir2020SI,HasLigRudBir2021NATUCOM,Lig2023phd}.
The different degree of robustness to the reality gap between \cho and \evsti had been only reported for \epucks.
Here, we show that similar results also apply to \mercators.
Demonstrative videos are available for download in the Supplementary material\footnote{\label{fn:supp}Supplementary material: https://iridia.ulb.ac.be/supp/IridiaSupp2023-001}.

\begin{figure}
    \centering
    \includegraphics[trim={10 5 8 5},clip,width=0.95\linewidth]{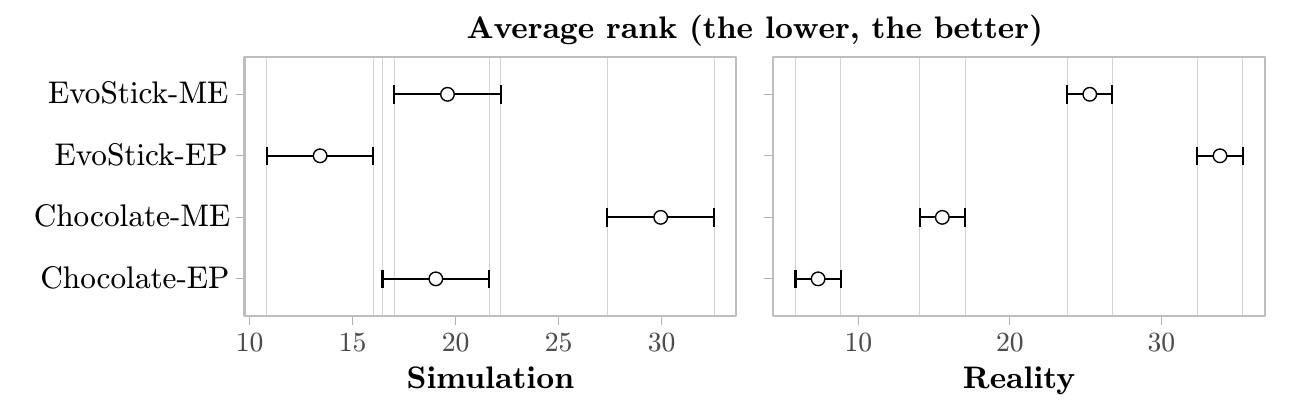}
    \caption{Comparison of control software produced by \evsti and \cho for \epucks~(EP) and \mercators~(ME). We aggregate the results obtained in the three missions using a Friedman test. For each method and platform, we present average ranks and $95\%$ confidence intervals. The lower, the better.}
    \label{fig:friedman-methods}
\end{figure}

To investigate the extent to which \cho and \evsti are transferable from the \epuck---the platform for which they were originally conceived---to the \mercator, we compare the performance of the control software they produce for the two platforms. 
We aggregate the performance across the three missions considered using a Friedman test~\cite{Con1999book}---see Figure~\ref{fig:friedman-methods}.
In simulation, the swarm of \epucks performs better than the one of \mercators, both for \cho and \evsti.
This result was expected, as the two design methods were conceived for the \epuck and were applied to \mercators without any adaptation.
In the experiments with the physical robots, the swarm of \epucks performs better than the one of \mercators when they execute control software produced by \cho.
On the other hand, the swarm of \mercators performs better than the \epucks when they execute control software produced by \evsti.
This result was unexpected and suggests that \evsti, although originally conceived for the \epucks, is more robust to the reality gap when adopted to design control software for \mercators.
This indicates that the effects of the reality gap are not only method-dependent but also platform-dependent.
By comparing the relative performance of \cho and \evsti, we can conclude that, under the experimental conditions considered in the study, \cho transfers better from \epucks to \mercators than \evsti---see Figure~\ref{fig:friedman-methods}. 

\begin{figure}
    \centering
    \includegraphics[trim={10 5 8 5},clip,width=0.95\linewidth]{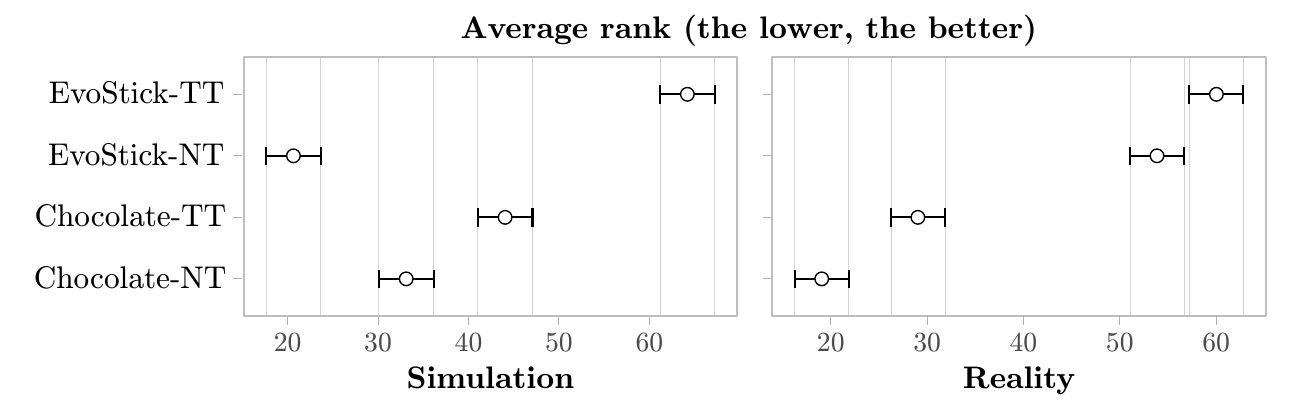}
    \caption{Comparison of control software produced by \evsti and \cho when assessed on the platform for which it has been produced~(NT, for \emph{not transferred}) and on the counterpart~(TT, for \emph{transferred})---regardless whether the design was performed for \epucks or \mercators. We aggregate the results obtained in the three missions using a Friedman test. We present average ranks and $95\%$ confidence intervals. The lower, the better.}
    \label{fig:friedman-ctrl}
\end{figure}

We also investigate the extent to which the control software automatically produced by \cho and \evsti can be transferred between \epucks and \mercators.
To this end, we transfer the control software produced for \epucks to \mercators, and \emph{vice versa}.
We assess the control software produced by the two methods both on the platform for which it has been produced and on the counterpart. 
Also in this case, we do this in simulation and reality.
We aggregate the performance across the three missions using a Friedman test---see Figure~\ref{fig:friedman-ctrl}.
When assessed in simulation, the control software produced by \evsti performs better than the one produced by \cho.
After transferring the control software, we observe that the one produced by \cho performs better than the one produced by \evsti---see Figure~\ref{fig:friedman-ctrl}.
The performance drop caused by transferring the control software is significantly larger for \evsti than for \cho.
We observe a rank inversion between the two design methods.
It is known that \evsti can achieve a better performance in simulation by \emph{overfitting} the design process to the simulated model of the \epuck~\cite{LigBir2020SI,Lig2023phd}.
We argue that this overfitting prevented a proper transfer of the control software to \mercator---as it happens when transferring control software between simulation and reality.
In the experiments with the physical robots, \cho and \evsti show a performance drop after transferring the control software between \epucks and \mercators.
However, the control software produced by \cho transfers better than the one produced by \evsti---see Figure~\ref{fig:friedman-ctrl}.

These preliminary results indicate that automatic design methods and the control software they produce can be transferred from one robot platform to another---provided that the two have equivalent sensing and actuation capabilities.
The best results are obtained by transferring a method, that is, by applying a method originally designed for a platform to another one---as opposed to transferring the control software it produces for the original platform to the other one. 
We will also investigate whether protocols to predict the robustness of design methods to the reality gap~\cite{Lig2023phd} can be used to predict the transferability of control software across platforms. 

\section*{Use of AI technologies in the writing process}
During the preparation of this work, the authors used OpenAI ChatGPT to proofread and edit for language issues and stylistic inconsistencies---see Supplementary material. After using this tool, the authors reviewed and edited the manuscript as needed and take full responsibility for the content.

\newpage

\bibliographystyle{IEEEtran} 
\IEEEtriggeratref{13}
\IEEEtriggercmd{\newpage}
\bibliography{demiurge-bib/definitions.bib,demiurge-bib/author.bib,demiurge-bib/address.bib,demiurge-bib/proceedings.bib,demiurge-bib/journal.bib,demiurge-bib/publisher.bib,demiurge-bib/series.bib,demiurge-bib/institution.bib,demiurge-bib/bibliography.bib,demiurge-bib/newbibliography.bib}

\end{document}